\documentclass[10pt,twocolumn,letterpaper]{article}

\usepackage{cvpr}              %
\usepackage[linesnumbered,ruled,vlined]{algorithm2e}
\usepackage{graphicx}
\usepackage{float}

\definecolor{cvprblue}{rgb}{0.21,0.49,0.74}
\usepackage[pagebackref,breaklinks,colorlinks,allcolors=cvprblue]{hyperref}
\usepackage{multirow} 

\def\paperID{7639} %
\def\confName{CVPR}
\def\confYear{2025}

\title{GaussianToken: An Effective Image Tokenizer with 2D Gaussian Splatting}

\author{Jiajun Dong\textsuperscript{\rm 1,}\footnotemark[1] \quad  
Chengkun Wang\textsuperscript{{\rm 2,}}\footnotemark[1]\thanks{Equal contribution. $\dagger$Project leader.} \quad 
Wenzhao Zheng\textsuperscript{{\rm 2,}}\footnotemark[2] \quad 
Lei Chen\textsuperscript{\rm 2} \quad 
Jiwen Lu\textsuperscript{\rm 2} \quad 
Yansong Tang\textsuperscript{\rm 1} \\
\textsuperscript{\rm 1}{Tsinghua Shenzhen International Graduate School, Tsinghua University, China}\\
\textsuperscript{\rm 2}{Department of Automation, Tsinghua University, China}\\
\tt \small \{dong-jj24,\ wck20\}@mails.tsinghua.edu.cn; 
\tt \small wenzhao.zheng@outlook.com}

\begin{document}
\maketitle
\begin{abstract}
Effective image tokenization is crucial for both multi-modal understanding and generation tasks due to the necessity of the alignment with discrete text data.
To this end, existing approaches utilize vector quantization (VQ) to project pixels onto a discrete codebook and reconstruct images from the discrete representation.
However, compared with the continuous latent space, the limited discrete codebook space significantly restrict the representational ability of these image tokenizers.
In this paper, we propose GaussianToken: An Effective Image Tokenizer with 2D Gaussian Splatting as a solution.
We first represent the encoded samples as multiple flexible featured 2D Gaussians characterized by positions, rotation angles, scaling factors, and feature coefficients.
We adopt the standard quantization for the Gaussian features and then concatenate the quantization results with the other intrinsic Gaussian parameters before the corresponding splatting operation and the subsequent decoding module.
In general, GaussianToken integrates the local influence of 2D Gaussian distribution into the discrete space and thus enhances the representation capability of the image tokenizer.
Competitive reconstruction performances on CIFAR, Mini-ImageNet, and ImageNet-1K demonstrate the effectiveness of our framework.
Our code is available at:~\url{https://github.com/ChrisDong-THU/GaussianToken}.

\end{abstract}
    
\section{Introduction}
Large Language Models (LLMs) have recently demonstrated dominance in natural language tasks because of the superior model capacity and scalability~\cite{achiam2023gpt,dubey2024llama3,yang2024qwen2}.
Additionally, a series of visual and multi-modal endeavors have attempted to exploit the auto-regressive architecture and pretrained knowledge from LLMs for vision-related tasks~\cite{liu2024visual,liu2024improved,tian2024visual}.
To accommodate the discrete input format of LLMs, they first tokenize images to obtain discrete visual tokens, and then perform text alignment and subsequent autoregressive predictions based on the task format.
Therefore, the effectiveness of image tokenizers directly determines the upper bound capabilities of the models.

Vector Quantization (VQ) is the prevalent image tokenization technique, balancing the requirements of image perception~\cite{bai2024sequential}, conditional image generation~\cite{tian2024visual,ramesh2021zero}, and multi-modal image understanding tasks~\cite{sun2024generative,wang2024emu3,lee2022autoregressive}.
Specifically, VQ-based strategies~\cite{van2017neural,esser2021taming,razavi2019generating} comprise a discrete codebook that encompasses a predetermined number of learnable vectors.
The encoded image features are aligned with the codebook vectors through similarity calculations to perform nearest-neighbor matching, which allows the image to be represented by discrete tokens derived from the codebook vectors.
Then a decoding module processes the discrete tokens to yield the reconstructed results in the RGB domain.
In addition, researchers have introduced a discriminator module to impose GAN-related constraints on the reconstructed images to enhance the authenticity and visual perception of images~\cite{esser2021taming}.
Nevertheless, compared to the continuous latent space of a naive VAE~\cite{kingma2013auto}, the size of the codebook space significantly limits the ability to model the distribution in the discrete space.
Consequently, this constraint imposes a decline in image reconstruction metrics and a redundancy in the training schedule.
Certain methods might augment the discrete space by increasing the codebook numbers or constructing lookup-free mappings~\cite{yu2023language,luo2024open}, while they inherently remain confined within the limited discrete space and necessitate a more gradual training process for convergence.

\begin{figure*}[t]
\centering
\includegraphics[width=0.99\textwidth]{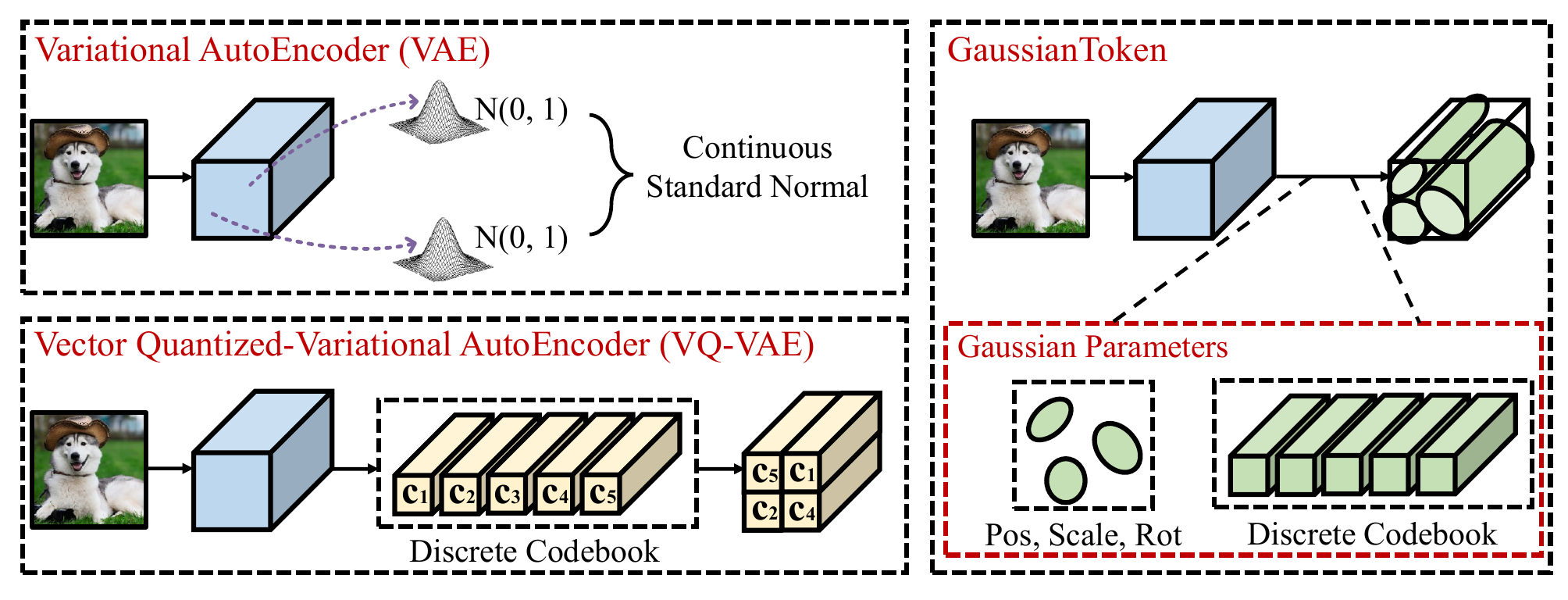}
\vspace{-4mm}
\caption{\textbf{Motivation of GaussianToken.} The Variational AutoEncoder (VAE) adopts a continuous space to represent image features, which fails to align with discrete modalities. In addition, Vector Quantization (VQ) approaches employ discrete codebooks for nearest-neighbor matching of image features, while the representation space is restricted to the codebook size. Differently, GaussianToken performs local adaptive learning on discrete features with continuous Gaussian parameters, enlarging the representation space under the discretized basis.
}
\label{fig:comparison}
\vspace{-6mm}
\end{figure*}

To address this, we propose an effective image tokenizer with 2D Gaussian Splatting, named GaussianToken, to enrich the codebook space for better modeling capabilities, as shown in Figure~\ref{fig:comparison}.
We propose a 2D Gaussian embedding module to parameterize the encoded image features into multiple Gaussian distributions. 
Each Gaussian distribution is characterized by its position, rotation angle, scaling factor, and feature coefficients, which is capable of learning adaptive independent parameters.
Subsequently, we propose to quantize the feature coefficients with a standard discrete codebook through nearest-neighbor matching, and concatenate the other parameters of the Gaussian distributions (position, rotation angle, and scaling factor) with the quantized results.
We then employ a 2D splatting module to project these combined Gaussian parameters back into the image feature space.
Ultimate procedures include a feature decoder that reconstruct the original images and a discriminator for further optimization.
Compared with conventional VQ-based approaches, GaussianToken presents a more flexible latent modeling strategy.
The feature representation is determined by the original discrete codebook, whereas the local feature attributes (such as position) are adaptively learned through 2D Gaussian distributions.
Consequently, GaussianToken constitutes a diverse combination within the discrete space via continuous Gaussian distributions, thereby expanding the representational capacity of the original discrete space.
We conduct extensive experiments on various datasets to access the effectiveness of the proposed GaussianToken, including CIFAR~\cite{krizhevsky2009learning}, Mini-ImageNet~\cite{vinyals2016matching}, and ImageNet-1K~\cite{russakovsky2015imagenet}.
Competitive reconstruction performances under similar settings demonstrate the superiority of our framework.

\section{Related Works}
\textbf{Image Tokenizer.}
Two typical image tokenizers include the straightforward patch embedding and the encoder-decoder architecture. 
Among them, the patch embedding operation in Vision Transformers (ViTs)~\cite{dosovitskiy2020image,touvron2021training,chu2021twins} aims at converting the original image into tokens that can be processed by the transformer structure. 
This approach is more commonly adopted in visual tasks such as image classification, which only require a global perception of the image.
On the contrary, image tokenizers with an encoder-decoder structure are adaptable to dense perception and generation task, which first proposed in VQ-VAE~\cite{van2017neural}.
They typically employ a discrete codebook to perform nearest-neighbor matching on image features and enhance the quality of reconstructed images with a lightweight discriminator~\cite{esser2021taming}.
To further enhance model performance, various approaches replace the original CNN backbone with a Vision Transformer (ViT) or a hybrid structure to strengthen feature extraction capabilities~\cite{esser2021taming,yu2021vector,villegas2022phenaki}. 
Other strategies focus on iterative and refined improvements in the codebook matching process~\cite{sun2024autoregressive,lee2022autoregressive}. 
In addition, MAGVIT-v2~\cite{luo2024open,yu2023language} substitutes the learnable codebook with a lookup-free format, which directly maps through numerical comparison, thereby expanding the equivalent number of codebooks to maximize the potential space.

However, these methods still compress the latent representation of images into a discrete space correlated with the size of the codebook, with features matched across spatial dimensions according to a single codebook, which directly constrains the representational power and reconstruction quality. 
To overcome the weakness, GaussianToken introduces a more adaptable quantization process through the use of 2D Gaussian Splatting, which adaptively learns spatially local information such as the position and scaling factor of the quantized features, thereby expanding the discrete space to achieve superior performance.

\textbf{Gaussian Splatting.}
Gaussian Splatting~\cite{kerbl20233d} first appeared in 3D scene reconstruction and is capable of addressing the real-time issues associated with model training and rendering in Neural Radiance Fields (NeRFs)~\cite{mildenhall2021nerf,pumarola2021d,yu2021pixelnerf}.
3D Gaussian Splatting researchers adopt explicit Gaussian ellipsoids and differentiable rasterization operations to achieve a high-quality and efficient rendering process, which facilitates both 3D realistic
scene reconstruction~\cite{kerbl20233d,luiten2023dynamic,cheng2024gaussianpro,yang2024deformable} and 3D scene editing~\cite{xie2024physgaussian,jiang2024gaussianshader,huang2024sc} tasks.
Additionally, several works have explored the applications of Gaussian Splatting in 2D image data~\cite{zhang2024gaussianimage,zhang2024image}.
For example, GaussianImage~\cite{zhang2024gaussianimage} introduced 2D Gaussian Splatting in the RGB domain for efficient image representation.
Image-GS~\cite{zhang2024image} further utilized 2D Gaussian Splatting for the efficient compression of individual images while achieving high-quality reconstruction results.

Nevertheless, these methods merely perform Gaussian learning on individual images and are unable to generalize to other samples for compression and reconstruction. 
Furthermore, Gaussian modeling represents a continuous latent space, which fails to integrate or align with discrete data modalities such as text. 
In this paper, our proposed GaussianToken introduces a local modeling process of 2D Gaussian Splatting within the VQ-VAEs.
GaussianToken enriches the image modeling capacity of the discrete space by adaptively learning the distribution of Gaussian features while ensuring the model’s generalization capability and the discrete nature of the latent space.

\section{Proposed Approach}
In this section, we provide a detailed exposition of our GaussianToken method based on 2D Gaussian Splatting.
We first provide a brief introduction of the VQ-VAE, whose discrete codebook space inherently constrains its representational ability.
To bridge this gap, we propose a novel quantization method, using featured 2D Gaussian (i.e., GaussianToken) as basic quantization units, to imbue the originally discrete space with a measure of continuity, 
while the feature coefficients of each unit remain discrete.
We then elaborate on the pivotal Gaussian Embedding framework, which enables the efficient learning of our proposed tokenized representation from raw image data.
Lastly, we provide an overall framework of GaussianToken and corresponding analysis of its effectiveness.

\subsection{Preliminaries: VQ-VAE}

VQ-VAE~\cite{van2017neural} is a unique type of variational autoencoder that adopts vector quantization to obtain a discrete latent representation. 
The key implementation lies in its latent embedding space $\mathcal{Q}=\{\mathbf{e}_1,\mathbf{e}_2,\dots,\mathbf{e}_N\}\in \mathbb{R}^{N\times D}$, where $N$ denotes the discrete latent space (a.k.a. the codebook size) and $D$ is the dimension of each latent embedding vector $\mathbf{z}_i$. 
Formally, given a high dimensional image $x\in \mathbb{R}^{H\times W \times C}$, the encoder $\mathcal{E}_\phi$ is first employed to produce the low dimensional latent representation $\hat{\mathbf{Z}}\in \mathbb{R}^{h\times w\times D}$. 
Note that $\hat{\mathbf{Z}}$ is a feature map, which at this stage, is comprised of a multitude of continuous latent variables.
Different from the typical autoencoder, VQ-VAE then passes $\hat{\mathbf{Z}}$ through a quantization module and equivalently transformes it into a set of indices representation, forming the quantized $\mathbf{Z}$. 
The index values correspond to the embeddings $\mathbf{z}_i$ of the original feature $\hat{\mathbf{z}}_i$ for $i \in [1, h\times w]$, determined by a nearest neighbor look-up using a shared codebook as follows:
\begin{equation}
\mathbf{z}_i=\mathbf{q}(\hat{\mathbf{z}}_i)=\mathbf{e}_k,~\text{where}~ k=\operatorname{argmin}_j\|\hat{\mathbf{z}}_i-\mathbf{e}_j\|_2,
\end{equation}
where $\mathbf{q}(\cdot)$ is the quantization operation. Ultimately, the decoder $\mathcal{D}_\theta$ inversely maps $\mathbf{Z}$ back to the image domain.
\begin{equation}
    \hat{x}=\mathcal{D}_\theta(\mathbf{Z})=\mathcal{D}_\theta(\mathbf{q}(\hat{\mathbf{Z}}))=\mathcal{D}_\theta(\mathbf{q}(\mathcal{E}_\phi(x))).
\end{equation}

The trainable components ($\mathcal{E}_\phi$, $\mathcal{D}_\theta$, and $\mathcal{Q}$) are optimized by minimizing the following objective:
\begin{equation}
    \mathcal{L}_\text{VQ-VAE}=\|x-\hat{x}\|_{2}^{2}+\|\mathrm{sg}[\mathbf{Z}]-\hat{\mathbf{Z}}\|_{2}^{2}+\beta\|\mathbf{Z}-\mathrm{sg}[\hat{\mathbf{Z}}]\|_{2}^{2},
\end{equation}
where $\mathrm{sg}[\cdot]$ denotes the stop-gradient operator and $\beta$ is a hyperparameter for loss balancing. 
VQ-VAE discretizes the continuous latent space into a codebook, where trained embedding vectors can be considered as foundational visual elements or intrinsic features that constitute the imagery. 
However, this characteristic implies that the representational capability of the codebook is highly sensitive to the codebook size and its utilization rate, which remains a core challenge that researchers in VQ-based generative models are striving to resolve.

\begin{figure*}[t]
\centering
\includegraphics[width=0.92\textwidth]{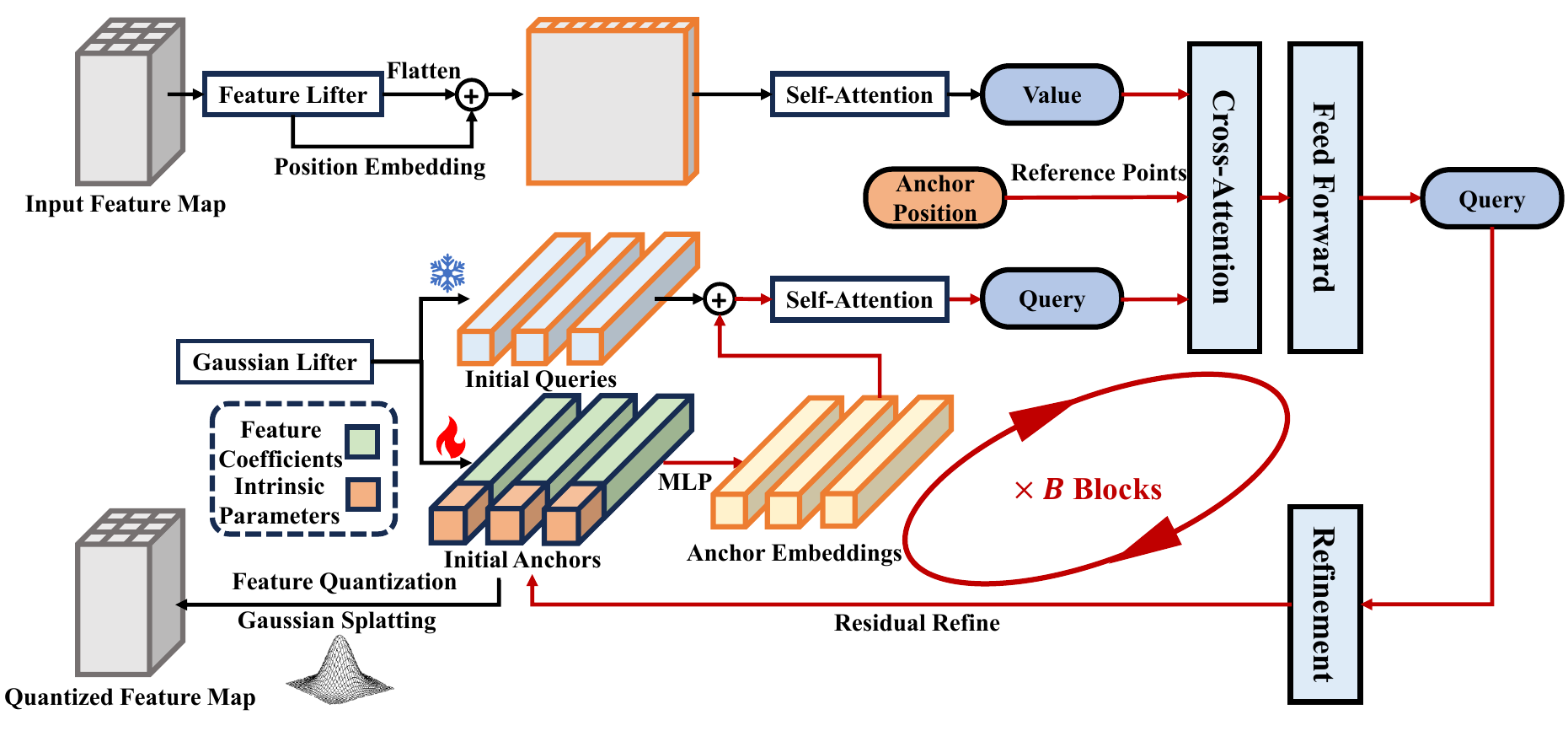}
\vspace{-5mm}
\caption{\textbf{Illustration of the proposed Gaussian Embedding module.} The information flow traverses through $B$ iterative blocks (marked in red arrow lines), with each block sequentially encompassing three primary operations: self-attention, cross-attention, and refinement, alternately updating the featured 2D Gaussian anchors under the guidance of visual information.
}
\label{fig:gsembed}
\vspace{-6mm}
\end{figure*}

\subsection{2D Gaussian Quantization}
Considering the above weakness, we propose a novel paradigm of feature quantization that introduces the concept of 2D Gaussian distribution. 
We extend the original quantization vector $\mathbf{z}_i$, which merely encompasses individual features, into a featured 2D Gaussian quantization unit $\mathbf{g}_k$ describing localized features within a certain region, rather than fixed grid.
Instead of directly replacing the quantization of feature vector $\hat{\mathbf{z}}_i$ by the corresponding quantized vector $\mathbf{z}_i$, we aggregate the contributions $\mathbf{c}_{ki}$ of all the quantization units $\mathbf{g}_k$ for $k\in[1,K]$ at position $\mathbf{p}_i=(x,y)$:
\begin{equation}
\mathbf{z}_i=\mathbf{q}^{\prime}(\hat{\mathbf{z}}_i)=\sum_{k=1}^{K}\mathbf{c}_{ki}, ~i\in[1, h\times w]
\label{eq-4}
\end{equation}
where K denotes the number of Gaussian units. 
Specifically, each featured 2D Gaussian unit is characterized by its position $\mathbf{\mu} \in\mathbb{R}^2$, covariance matrix $\mathbf{\Sigma}\in\mathbb{R}^{2\times2}$ and additional feature coefficient $\mathbf{\zeta}\in\mathbb{R}^D$.
Therefore, the contribution $\mathbf{c}_{ki}$ can be formulated with the probability $\mathbf{\pi}_{ki}$ of 2D Gaussian distribution as follows:
\begin{equation}
    \mathbf{c}_{ki}=\mathbf{\pi}_{ki}\mathbf{\zeta}_k=\exp \left(-\frac{1}{2}(\mathbf{p}_i-\mathbf{\mu}_k)^T \mathbf{\Sigma}^{-1}(\mathbf{p}_i-\mathbf{\mu}_k)\right) \mathbf{\zeta}_k.
\end{equation}

Since the covariance matrix $\mathbf{\Sigma}$ of a Gaussian distribution must be positive semi-definite, it is necessary to ensure valid matrix values during the numerical optimization process.
Thus we opt to refine the factorized representation of the covariance matrix with the product of a rotation matrix $\mathbf{R}\in\mathbb{R}^{2\times2}$ and scaling matrix $\mathbf{S}\in\mathbb{R}^{2\times2}$:
\begin{equation}
    \mathbf{\Sigma}=(\mathbf{R}\mathbf{S})(\mathbf{R}\mathbf{S})^{T},
\end{equation}
where $\mathbf{R}$ and $\mathbf{S}$ derive from on the rotation angle $\theta\in[0,\pi]$ and scaling factors $\mathbf{s}\in\mathbb{R}^2$ as follows:
\begin{equation}
    \mathbf{R}=\left[ \begin{array}{cc}
\cos(\theta) & -\sin(\theta) \\ \sin(\theta) & \cos(\theta) \end{array} \right], ~\mathbf{S}=\left[ \begin{array}{cc}
s_1 & 0 \\ 0 & s_2 \end{array} \right].
\end{equation}

We employ the framework detailed in Section\ref{sec3.2} to transform image data from the RGB domain into a set of the aforementioned featured 2D Gaussian units:
\begin{equation}
    \mathcal{G}=\{\mathbf{g}_k\in \mathbb{R}^d|k=1,2,\dots,K\}, 
\end{equation}
where $\mathbf{g}_k:=\mathbf{g}_k(\mathbf{\mu}_k,\theta_k,\mathbf{s}_k,\mathbf{\zeta}_k)$ encapsulates the essence of one unit, with \(d = 5 + D\) denoting its total parameter dimension. 
We first quantize the feature coefficients $\mathbf{\zeta}_k, k\in[1,K]$ following the same approach as in the conventional VQ-VAE, while maintaining the continuity of their positions and covariance matrices.
Under such circumstances, the model is capable to optimize the units to any position and scale within the feature map, which allows the 2D Gaussian to adaptively allocate computation and storage resources according to the region complexities. 
The quantized units are subsequently aggregated into the quantized feature map with ultra-fast rendering via a 2D Gaussian Splatting function efficiently implemented with CUDA.

\begin{figure*}[t]
\centering
\includegraphics[width=0.95\textwidth]{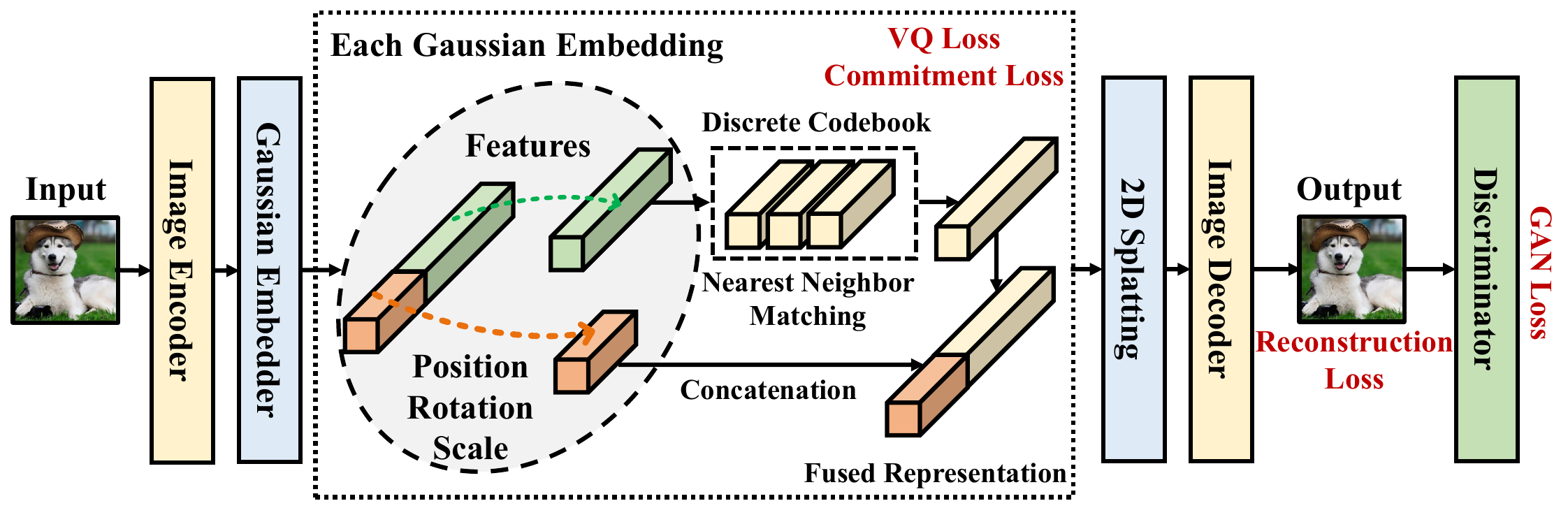}
\caption{\textbf{The overall framework of GaussianToken.} We process image features using the Gaussian Embedder, obtaining intrinsic spatial parameters (position, rotation angle, and scaling factor) as well as feature coefficients. We perform nearest neighbor matching with a discrete codebook for the feature coefficients to achieve the quantization process and concatenate the quantized results with the spatial parameters. Subsequently, we adopt a 2D splatting module and an image decoder to generate the corresponding reconstructed image. Finally, we impose quality constraints on the reconstructed image through an additional discriminator. The overall constraints include the image reconstruction loss, the quantization-related VQ loss and commitment loss, as well as the GAN loss for image quality.
}
\label{fig:framework}
\vspace{-6mm}
\end{figure*}

\subsection{Gaussian Embedding}
\label{sec3.2}
Building upon the 2D Gaussian quantization discussed above, we further propose a Gaussian Embedding module to learn meaningful Gaussian representations leveraging image features. 
As depicted in Figure \ref{fig:gsembed}, our pipeline commences with two primary information carriers: the pristine image features and the 2D Gaussian objects.
Essentially, we adopt the Gaussian Embedding module as a conduit for information exchange, which keeps refining the learned properties through a series of operations.
Below, we delve into the intricacies and underpinnings of the relevant operations.

\textbf{Lifter Module.}
As a preparatory step for subsequent modules, the Lifter Module transforms the two primary information carriers into unified vectors respectively.
To adapt it for attention architecture, the Feature Lifter flattens the feature map $\hat{\mathbf{Z}}$ from encoder $\mathcal{E}$ into a sequence of features $\hat{\mathbf{S}}\in \mathbb{R}^{(h\times w)\times D^{\prime}}$.
Note that $D^{\prime}$ is significantly larger than the channel dimension $D$ of the desired quantized feature map $\mathbf{Z}$ so as to preserve more image information for Gaussian representations learning.
Additionally, we also concatenate the sequence with a cosine position embedding, thereby equipping the model with the capability to discern the order.
On the other hand, the Gaussian Lifter initiates the properties of the featured 2d Gaussian anchors $\mathbf{G}$ and their associated high-dimensional queries $\mathbf{Q}$. 
Since each anchor $\mathbf{g}_k$ is refined in the form of target Gaussian parameters, we maintain a multi-layer linear perceptron (MLP) to get the embedding feature $\hat{\mathbf{g}}_k\in\mathbb{R}^{D^{\prime}}$ to ensure seamless interaction with the query.
Eventually, we obtain the unified representation of the two entities in the latent space with an embedding dimension of $D^{\prime}$.

\textbf{Self-attention.}
We employ the self-attention layers on both visual features sequence $\hat{\mathbf{S}}$ for further compressing image information and queries $\mathbf{Q}$ for incorporating interactions among 2D Gaussian anchors.
Note that we substitute the Deformable attention (DA)~\cite{zhudeformable} for the Transformer attention~\cite{kim2018attention} when processing the visual features sequence, thereby mitigating the high computational complexity of \( O(N^2) \) and the challenge of handling high-resolution features, which we will elaborate on in the following module.

\textbf{Cross-attention.}
The randomly initialized 2D Gaussian objects extract visual information within the cross-attention (CA) module, which is also based on DA.
Specifically, for the 2D Gaussian anchor $\mathbf{g}_k$, we generate a series of offsets $\Delta\mathbf{\mu}_{ki}$. 
We further derive a series of reference points $\mathcal{R}_k=\{\mathbf{\mu}_k+\Delta\mathbf{\mu}_{ki} | i=1,\dots,R\}$ through combining $\Delta\mathbf{\mu}_{ki}$ with the position of the anchor $\mathbf{\mu}_k$.
Then we calculate the corresponding attention weights $\mathbf{A}_{ki}$, which are subsequently used for weighted summation of the \emph{values}:
\begin{equation}
    \operatorname{CA}\left(\mathbf{g}_k, \mathbf{q}_k, \boldsymbol{x}\right)=\sum_{i=1}^R A_{k i} \cdot \boldsymbol{W}\boldsymbol{x}\left(\mathbf{\mu}_k+\Delta\mathbf{\mu}_{ki}\right),
\end{equation}
where $\boldsymbol{x}$ denotes the self-encoded image features and $\boldsymbol{W}$ represents the linear transformation matrix that projects them to the \emph{values}.
The scalar attention weight $\mathbf{A}_{ki}$ lies in the range $[0, 1]$, normalized by $\Sigma_{i=1}^{R}\mathbf{A}_{ki}=1$. 
Both $\Delta\mathbf{\mu}_{ki}$ and $\mathbf{A}_{ki}$ are obtained via linear projection over the query feature $\mathbf{q}_k$ corresponding to the anchor $\mathbf{g}_k$. 
In implementation, we merge the embedding feature $\hat{\mathbf{g}}_k$ of the anchor $\mathbf{g}_k$ with the query feature $\mathbf{q}_k$ prior to the module input, which enhances the connection between them.

\textbf{Refinement.}
We employ the refinement module to rectify the properties of anchors $\mathbf{G}$ with guidance of the query result $\hat{\mathbf{Q}}$ from the preceding cross-attention module. 
To elaborate, we first decode the property adjustment quantities $\Delta\mathbf{g}$ for anchor $\mathbf{g}\in\mathbf{G}$ from $\hat{\mathbf{Q}}$ via a MLP:
\begin{equation}
    \Delta\mathbf{g}=\{\Delta\mathbf{\mu},\Delta\theta,\Delta\mathbf{s},\Delta\mathbb{\zeta}\}=\operatorname{MLP}(\hat{\mathbf{Q}}).
\end{equation}
In the specific adjustment strategy, we employ a more rational approach that we directly replace the variables requiring rapid adjustment in $\mathbf{g}$ with $\Delta\mathbf{g}$, including the rotation angle $\theta$, the scale factors $\mathbf{s}$ and the feature coefficient $\mathbf{\zeta}$, except for the position $\mathbf{\mu}$. 
Considering that $\mathbf{\mu}$ determines the region within the feature map affected by the anchor, frequent replacement to $\mathbf{\mu}$ could disrupt the optimization of all the other properties, leading to instability in the training process.
Instead, we refine $\mathbf{\mu}$ of the anchor by adding a residual adjustment quantity $\Delta\mathbf{\mu}$ to it:
\begin{equation}
    \mathbf{g}_{new}=\{\mathbf{\mu}+\Delta\mathbf{\mu},\Delta\theta,\Delta\mathbf{s},\Delta\mathbf{\zeta}\}.
\end{equation}

\begin{table*}[t]
    \centering
    \caption{Image reconstruction results on CIFAR, Mini-ImageNet, and ImageNet-1K. ($\dag$ denotes our reproduced results.)}
    \vspace{-2mm}
    \renewcommand\arraystretch{0.88}
    \setlength\tabcolsep{3.4pt}
    \begin{tabular}{lcccccccccc}
    \toprule
    \multirow{2}{*}{Method} & \multirow{2}{*}{Dataset} & Image & \multirow{2}{*}{Tokens} & \multirow{2}{*}{Ratio} & Embedding & Codebook & \multirow{2}{*}{Epoch} & \multirow{2}{*}{rFID$\downarrow$} & \multirow{2}{*}{PSNR$\uparrow$} & \multirow{2}{*}{SSIM$\uparrow$} \\
     & & Size & & & Dimension & Size & & & & \\
    \midrule
    VQ-VAE~\cite{van2017neural} & \multirow{6}{*}{CIFAR} & 32$\times$32 & 8$\times$8 & 4 & 256 & 1024 & 500 & 39.67 & - & 0.86 \\
    HVQ-VAE~\cite{williams2020hierarchical} &  & 32$\times$32 & 8$\times$8 & 4 & 256 & 1024 & 500 & 41.08 & - & 0.86 \\
    SQ-VAE~\cite{takida2022sq} &  & 32$\times$32 & 8$\times$8 & 4 & 256 & 1024 & 500 & 37.92 & - & 0.88 \\
    VQGAN$\dag$~\cite{esser2021taming} &  & 32$\times$32 & 8$\times$8 & 4 & 4 & 1024 & 30 & 27.00 & 23.59 & 0.81 \\
    CVQ-VAE~\cite{zheng2023online} &  & 32$\times$32 & 8$\times$8 & 4 & 256 & 1024 & 500 & 24.73 & - & 0.90 \\
    \textbf{GaussianToken} &  & 32$\times$32 & 64 & 4 & 3 & 1024 & 30 & \textbf{12.94} & \textbf{25.84} & \textbf{0.90} \\
    \midrule
    VQGAN$\dag$~\cite{esser2021taming} & \multirow{2}{*}{Mini-ImageNet} & 256$\times$256 & 16$\times$16 & 16 & 8 & 1024 & 30 & 32.31 & 18.71 & 0.50 \\
    \textbf{GaussianToken} &  & 256$\times$256 & 256 & 16 & 8 & 1024 & 30 & \textbf{12.18} & \textbf{19.84} & \textbf{0.57} \\
    \midrule
    VQ-VAE-2~\cite{razavi2019generating} & \multirow{6}{*}{ImageNet-1K} & 256$\times$256 & 32$\times$32 & 8 & 256 & 512 & - & $\sim$ 10 & - & - \\
    VQGAN~\cite{esser2021taming} &  & 256$\times$256 & 16$\times$16 & 16 & 256 & 1024 & 60 & 7.94 & 19.40 & - \\
    SD-VQGAN~\cite{rombach2022high} &  & 256$\times$256 & 16$\times$16 & 16 & 256 & 16384 & - & 5.15 & - & - \\
    MaskGIT~\cite{chang2022maskgit} &  & 256$\times$256 & 16$\times$16 & 16 & 256 & 1024 & 50 & 2.28 & - & - \\
    LlamaGen~\cite{sun2024autoregressive} &  & 256$\times$256 & 16$\times$16 & 16 & 256 & 16384 & 40 & 2.19 & \textbf{20.79} & - \\
    \textbf{GaussianToken} &  & 256$\times$256 & 256 & 16 & 8 & 1024 & 20 & \textbf{1.61} & 20.68 & \textbf{0.58} \\
    \bottomrule
    \end{tabular}
    \vspace{-5mm}
    \label{tab:main result}
\end{table*}

\subsection{GaussianToken}
We present the overall framework of our GaussianToken, as illustrated in Figure \ref{fig:framework}.
Subsequently, We elucidate the advantages of GaussianToken over previous methods from the fundamental structural design.

\textbf{Analysis.}
Firstly, the sparsity of the 2D Gaussians promotes an efficient representation capability. 
Capitalizing on the local influence of 2D Gaussian distribution, each GaussianToken can affect all features $\mathbf{z}$ within a surrounding region.
Therefore, with an equivalent quantity of tokens, GaussianToken constructs a more flexible discrete representation space compared to VQ-VAE~\cite{van2017neural}.
Practically, we might achieve superior reconstruction performances with a smaller embedding dimension.

In addition, GaussianToken presents an effectively accelerated convergence rate during training.
We define the valid coverage region ($\mathbf{c}_{ki}\neq\mathbf{0}$) of the 2D Gaussian distribution of unit $\mathbf{g}_k$ as $\Omega_k$. 
The error propagated to $\mathbf{g}_k$ during the backpropagation optimization can be calculated as:
\begin{equation}
    \mathcal{L}(\mathbf{g}_k)=\sum_{\mathbf{p}_i\in \Omega_k}\mathcal{L}(\mathbf{z}_i)\cdot\mathbf{\pi}_{ki} ,
\end{equation}
where the summation is numerically scaled by the distribution probabilities $\mathbf{\pi}_{ki}$. 
This reveals that our method can efficiently utilize regional error sums to optimize each unit $\mathbf{g}_k$, leading to a smoother and faster model convergence. 
Our experimental results fully substantiate this perspective.

Lastly, our semi-discrete representation space of the featured 2D Gaussian codebook is superior to that in VQ-VAE~\cite{van2017neural} for the potential downstream generation task. 
Specifically, assuming both methods learn a fixed number of $N$ embedding vectors codebook at the image tokenization stage, VQ-VAE typically models the prior distribution of the encoded indices matrix $\mathcal{M}\in\mathbb{R}^{h\times w}$ in an autoregressive manner to achieve realistic image synthesis. 
Differently, GaussianToken models the prior distribution of an unordered encoded indices set $\mathcal{S}\in\mathbb{R}^{K}$ and then infers the remaining intrinsic parameters of GaussianTokens from the indexed feature vectors.
Ideally, the autoregressive model of VQ-VAE generates a indices matrix $\mathcal{M}^{\prime}$ with a maximum possible combination count of $M=(h\times w)^N$, which is equivalent to the diversity of the generated images.
Nevertheless, considering the specific meaning of the tokenized representation in the codebook, this number is significantly reduced.
By comparison, as indicated in \eqref{eq-4}, each feature $\mathbf{z}_i$ in our generated feature map varies with the contributions of all surrounding related GaussianTokens, resulting in a diversity of feature map that far exceeds $M$.
Note that this paper focuses on the design of the image tokenizer without including specific experiments on generation tasks, which we regards as the future work.

\textbf{Training.}
GaussianToken can be easily integrated into the existing visual tokenizers with a mere alteration to the quantization process. 
To benchmark against the state-of-the-art VQ-VAE methods, we deploy our approach by simply plugging the Gaussian Embedding module into the VQGAN~\cite{esser2021taming}.
The overall loss consists of the reconstruction loss $\mathcal{L}_\text{rec}$, the commitment loss $\mathcal{L}_\text{c}$ and the additional GAN loss $\mathcal{L}_\text{GAN}$, which is formulated as follows:
\begin{equation}
    \mathcal{L}_\text{VQ-GAN}=\mathcal{L}_\text{rec}+\alpha\mathcal{L}_\text{c}+\beta\mathcal{L}_\text{GAN},
\end{equation}
where $\alpha$ and $\beta$ balance the three terms. 

\begin{table*}[t]
\centering
\begin{minipage}{0.33\textwidth}
    \centering
    \vspace{0.4mm}
    \caption{Effect of the embedding dimension.}
    \vspace{-3mm}
    \renewcommand\arraystretch{0.88}
    \setlength\tabcolsep{4.5pt}
    \begin{tabular}{c ccc}
    \toprule
    Embedding & \multirow{2}{*}{rFID$\downarrow$} & \multirow{2}{*}{PSNR$\uparrow$} & \multirow{2}{*}{SSIM$\uparrow$} \\
    Dimension &  &  & \\
    \midrule
    2 & 16.34 & 24.67 & 0.86 \\
    3 & \textbf{12.94} & \textbf{25.84} & \textbf{0.90} \\
    4 & 13.89 & 25.50 & 0.88 \\
    8 & 13.86 & 25.08 & 0.88 \\
    \bottomrule
    \end{tabular}%
    \vspace{-4mm}
    \label{tab:embed_dim}%
    \end{minipage}
\hfill
\begin{minipage}{0.33\textwidth}
\centering
    \caption{Effect of the codebook size.}
    \vspace{-3mm}
    \renewcommand\arraystretch{0.88}
    \setlength\tabcolsep{4.5pt}
    \begin{tabular}{c ccc}
    \toprule
    Codebook & \multirow{2}{*}{rFID$\downarrow$} & \multirow{2}{*}{PSNR$\uparrow$} & \multirow{2}{*}{SSIM$\uparrow$} \\
    Size &  &  & \\
    \midrule
    512 & 14.56 & 24.86 & 0.87 \\
    1024 & 13.89 & 25.50 & 0.88 \\
    2048 & \textbf{12.96} & 25.97 & \textbf{0.89} \\
    16384 & 13.03 & \textbf{26.16} & 0.89 \\
    \bottomrule
    \end{tabular}%
    \vspace{-4mm}
    \label{tab:codebook size}%
    \end{minipage}
\hfill
\hspace{-0.3cm}
\begin{minipage}{0.33\textwidth}
\centering
    \caption{Effect of the Gaussian number.}
    \vspace{-3mm}
    \renewcommand\arraystretch{0.88}
    \setlength\tabcolsep{4.5pt}
    \begin{tabular}{c ccc}
    \toprule
    Gaussian & \multirow{2}{*}{rFID$\downarrow$} & \multirow{2}{*}{PSNR$\uparrow$} & \multirow{2}{*}{SSIM$\uparrow$} \\
    Number &  &  & \\
    \midrule
    32 & 20.09 & 23.01 & 0.80 \\
    64 & 13.89 & 25.50 & 0.88 \\
    128 & 10.14 & 26.98 & 0.92 \\
    256 & \textbf{7.19} & \textbf{28.23} & \textbf{0.95} \\
    \bottomrule
    \end{tabular}%
    \vspace{-4mm}
    \label{tab:Gaussian Number}%
\end{minipage}
\end{table*}

\begin{figure*}[t]
\centering
\vspace{1mm}
\includegraphics[width=0.99\textwidth, trim=0cm 0cm 0cm 0.8cm, clip]{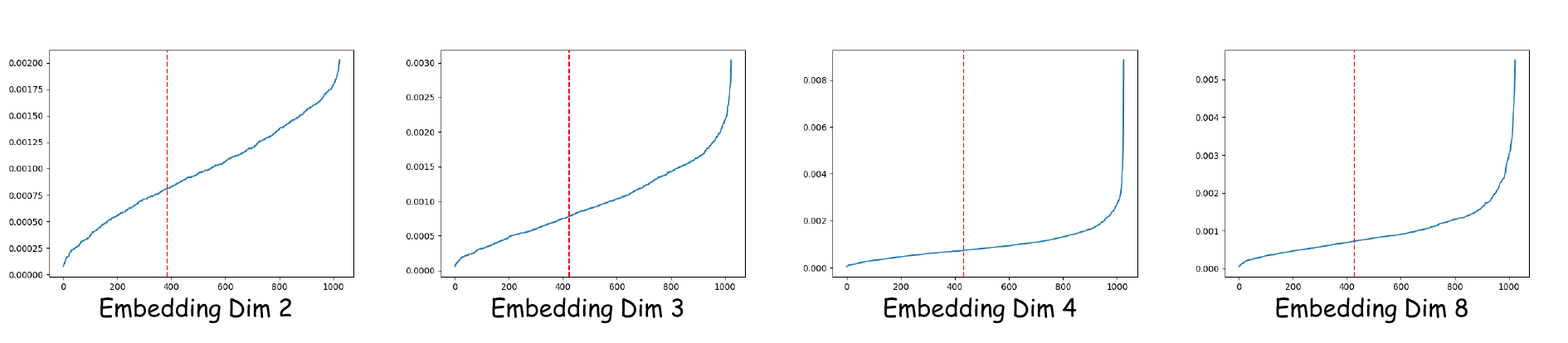}
\vspace{-4mm}
\caption{\textbf{Frequency of vector usage in the codebook.} The red dashed line represents the 20\% frequency cumulative line. 
}
\label{fig:codebook_usage}
\vspace{-5mm}
\end{figure*}

\section{Experiments}
In this section, we conducted extensive experiments to verify the effectiveness of the proposed GaussianToken. 
We performed image reconstruction tasks on the CIFAR, Mini-ImageNet, and ImageNet-1K datasets, respectively.
Additionally, we provided in-depth ablation studies for corresponding analysis and reconstructed visualization results for more intuitive comparisons.
All our experiments were conducted on 8 RTX 3090 GPUs.

\subsection{Datasets}
The CIFAR~\cite{krizhevsky2009learning} dataset can be divided into CIFAR-10 and CIFAR-100 based on the categories, while we ignore the category factor for image reconstruction.
The entire dataset contains 60,000 samples with a resolution of $32\times 32$, with the training set comprising 50,000 images and the test set containing 10,000 images.
Mini-ImageNet~\cite{vinyals2016matching} is a carefully curated few-shot learning dataset derived from ImageNet-1K~\cite{russakovsky2015imagenet}, featuring a higher spatial resolution compared to CIFAR. 
Mini-ImageNet encompasses 100 distinct categories, each with 600 images. 
We employ 48,000 images for training and the remaining 12,000 images for testing. 
Additionally, ImageNet-1K comprises 1,000 categories, with the training set containing over 1,280,000 samples, while the test set consists of 5,000 images.

\subsection{Implementation Details}
We conveniently implemented GaussianToken by adding the Gaussian Embedding module after the encoder of the original VQGAN~\cite{esser2021taming} baseline while maintaining other structures and hyper-parameters. 
We set the resolutions of input images as $32 \times 32$ for CIFAR and $256\times 256$ for Mini-ImageNet and ImageNet-1K. 
The downsampling ratios for the three datasets are 4, 16, and 16, respectively.
We fixed the embedding dimension to 4 for CIFAR and 8 for Mini-ImageNet and ImageNet-1K.
We used a codebook size of 1024 for both CIFAR and Mini-ImageNet, while we set the size to 2048 for ImageNet-1K.
We adopted $B=3$ transformer blocks in the Gaussian Embedding module as the default setting to refine the properties of 2D Gaussians. 
We respectively employed two Adam~\cite{diederik2014adam} optimizers with the same settings for the discriminator and the other GaussianToken model structures.
We set the base learning rate to $1\times10^{-4}, \beta_1=0.5, \beta_2= 0.9$, and the weight decay to $0$.
We adopted a cosine schedule with the warmup epoch of 1.
We respectively trained our models for 30, 30, and 20 epochs for the three datasets considering the training time.

\begin{figure*}[t]
    \centering
    \includegraphics[width=0.99\textwidth, trim=2cm 1cm 1.5cm 1cm, clip]{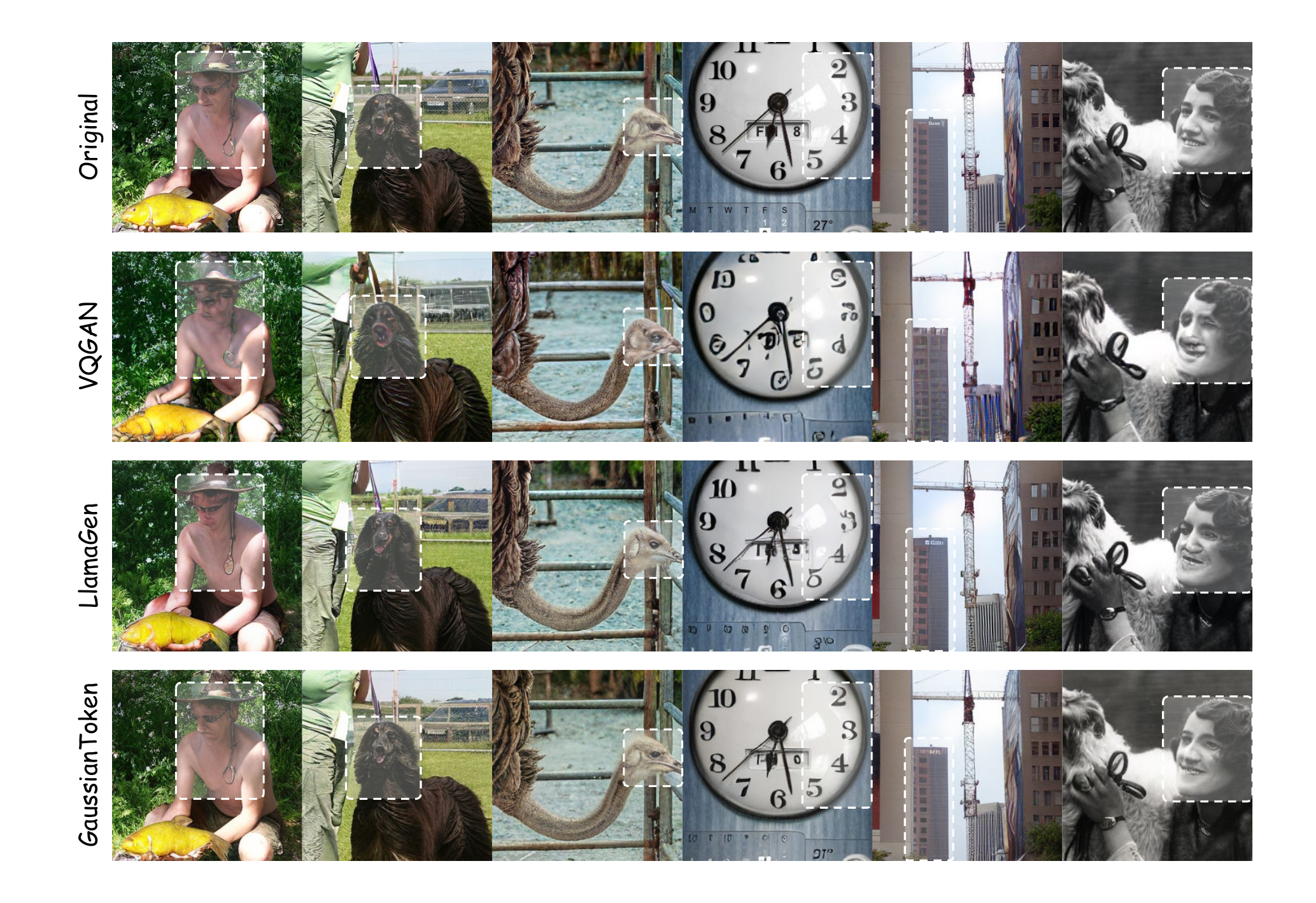}
    \caption{\textbf{Reconstruction comparison across different tokenizers.} Compared to the baseline VQGAN~\cite{esser2021taming} and LlamaGen~\cite{sun2024autoregressive}, our proposed model substantially enhances reconstruction quality (highlighted in the white box) while utilizing the same number of image tokens (256 tokens for a 256x256 resolution).}
    \label{fig:visualization}
    \vspace{-6mm}
\end{figure*}

\subsection{Main Results}
We compared our GaussianToken framework with existing methods under similar settings across three datasets.
We ensured consistent spatial resolutions (number of tokens) and downsampling ratios.
The comparison results are presented in Table~\ref{tab:main result}.
We observe that GaussianToken demonstrates significant efficiency compared to the baselines.
For example, on the CIFAR dataset, GaussianToken with an embedding dimension of 4 and a training epoch of 30 outperforms VQGAN by
14.01 in rFID.
In addition, GaussianToken can also achieve superior results on the Mini-ImageNet and ImageNet-1K datasets with less training time and a smaller embedding dimension compared to the comparison approaches.
In particular, compared to LlamaGen~\cite{sun2024autoregressive}, GaussianToken achieves a better reconstruction performance on ImageNet-1K with a smaller embedding dimension (8 vs. 256), a smaller codebook size (1024 vs. 16384), and fewer training epochs (20 vs. 40), resulting in an FID of 1.67 vs. 2.19.
This is because GaussianToken leverages Gaussian parameters for adaptive optimization of local features, resulting in a discrete space with stronger representational capabilities under similar conditions.

\subsection{Ablation Studies}
To further understand our GaussianToken, we conducted various ablation studies including embedding dimension, codebook usage, codebook size, and Gaussian number to verify its effectiveness on the CIFAR dataset.

\textbf{Embedding Dimension and Codebook Usage.}
A significant advantage of GaussianToken is its ability to achieve efficient reconstruction with a relatively smaller embedding dimension.
Therefore, we respectively used 2, 3, 4, and 8 as the embedding dimension, shown in Table \ref{tab:embed_dim}.
We find that the reconstruction performance of the model first increases and then decreases with the expansion of the embedding dimension.
In addition, GaussianToken reaches the best performance when the dimension equals to 3.
We further provided the codebook usage corresponding to different embedding dimensions in Figure \ref{fig:codebook_usage}.
We observe that a smaller embedding dimension corresponds to a higher codebook utilization rate.
Consequently, GaussianToken maintains a high codebook utilization rate without fully compressing the discrete space when the embedding dimension is equal to 3, thereby achieving optimal performance.

\textbf{Codebook Size.}
We verified the reconstruction performance of GaussianToken under codebook sizes of 512, 1024, 2048 and 16384, respectively.
The comparison results are illustrated in Table \ref{tab:codebook size}.
We observe that GaussianToken is not sensitive to changes in codebook size, which is because we adaptively perform local optimization on the limited codebook space to inherently enhance the representational capability of the discrete codebook.
GaussianToken achieves optimal performance on two main metrics when the codebook size is 2048.
As the codebook size increases, the utilization of the codebook decreases and finally falling below 50\% when the size reaches 16,384. 
While codebook collapse is a major drawback of the traditional vector quantization method in VQGAN, this technical issue is not the focus of this paper and will not be discussed further.
To rigorously validate our method, we set the codebook size to 1024 as the main experimental setup.

\textbf{Gaussian Number.}
The Gaussian number directly affects the modeling capability, which we adjust from 16 to 128 to compare the reconstruction performance in Table \ref{tab:Gaussian Number}.
We observe that the performance progressively improves as the number of Gaussian components increases, albeit with a diminishing rate of enhancement.
Specially, the gaussian number of 64 demonstrates a relatively superior result without an excessive computational cost of the model.

\subsection{Visualization}
We visualized the image reconstruction results of GaussianToken on ImageNet-1K in Figure \ref{fig:visualization}.
We observe that GaussianToken is capable of reconstructing images of high quality and exhibits realistic details and textures compared with original samples.

\section{Conclusion}
In this paper, we have proposed GaussianToken as an effective image tokenizer with 2D Gaussian Splatting.
We have adaptively learned the local quantization feature ranges through Gaussian positions, rotation angles, and scaling factors, which enhances the representational capability of the latent discrete space.
We have further rendered all Gaussian parameters and obtained the final reconstruction results via an image decoder.
We have validated the effectiveness of the proposed GaussianToken on the image reconstruction task and provided corresponding ablation analysis for performance comparisons.

\textbf{Limitations.}
GaussianToken focuses solely on the design of the discrete codebook space, dedicated to enhancing the corresponding representation capabilities, while the evaluation on downstream tasks such as image generation remains unexplored.
Future work will involve comprehensive comparison experiments for image generation.

{
    \small
    \bibliographystyle{ieeenat_fullname}
    \bibliography{main}

\begin{thebibliography}{46}
\providecommand{\natexlab}[1]{#1}
\providecommand{\url}[1]{\texttt{#1}}
\expandafter\ifx\csname urlstyle\endcsname\relax
  \providecommand{\doi}[1]{doi: #1}\else
  \providecommand{\doi}{doi: \begingroup \urlstyle{rm}\Url}\fi

\bibitem[Achiam et~al.(2023)Achiam, Adler, Agarwal, Ahmad, Akkaya, Aleman, Almeida, Altenschmidt, Altman, Anadkat, et~al.]{achiam2023gpt}
Josh Achiam, Steven Adler, Sandhini Agarwal, Lama Ahmad, Ilge Akkaya, Florencia~Leoni Aleman, Diogo Almeida, Janko Altenschmidt, Sam Altman, Shyamal Anadkat, et~al.
\newblock Gpt-4 technical report.
\newblock \emph{arXiv preprint arXiv:2303.08774}, 2023.

\bibitem[Bai et~al.(2024)Bai, Geng, Mangalam, Bar, Yuille, Darrell, Malik, and Efros]{bai2024sequential}
Yutong Bai, Xinyang Geng, Karttikeya Mangalam, Amir Bar, Alan~L Yuille, Trevor Darrell, Jitendra Malik, and Alexei~A Efros.
\newblock Sequential modeling enables scalable learning for large vision models.
\newblock In \emph{CVPR}, pages 22861--22872, 2024.

\bibitem[Chang et~al.(2022)Chang, Zhang, Jiang, Liu, and Freeman]{chang2022maskgit}
Huiwen Chang, Han Zhang, Lu Jiang, Ce Liu, and William~T Freeman.
\newblock Maskgit: Masked generative image transformer.
\newblock In \emph{CVPR}, pages 11315--11325, 2022.

\bibitem[Cheng et~al.(2024)Cheng, Long, Yang, Yao, Yin, Ma, Wang, and Chen]{cheng2024gaussianpro}
Kai Cheng, Xiaoxiao Long, Kaizhi Yang, Yao Yao, Wei Yin, Yuexin Ma, Wenping Wang, and Xuejin Chen.
\newblock Gaussianpro: 3d gaussian splatting with progressive propagation.
\newblock In \emph{ICML}, 2024.

\bibitem[Chu et~al.(2021)Chu, Tian, Wang, Zhang, Ren, Wei, Xia, and Shen]{chu2021twins}
Xiangxiang Chu, Zhi Tian, Yuqing Wang, Bo Zhang, Haibing Ren, Xiaolin Wei, Huaxia Xia, and Chunhua Shen.
\newblock Twins: Revisiting the design of spatial attention in vision transformers.
\newblock In \emph{NeurIPS}, 2021.

\bibitem[Diederik(2014)]{diederik2014adam}
P~Kingma Diederik.
\newblock Adam: A method for stochastic optimization.
\newblock \emph{(No Title)}, 2014.

\bibitem[Dosovitskiy et~al.(2020)Dosovitskiy, Beyer, Kolesnikov, Weissenborn, Zhai, Unterthiner, Dehghani, Minderer, Heigold, Gelly, et~al.]{dosovitskiy2020image}
Alexey Dosovitskiy, Lucas Beyer, Alexander Kolesnikov, Dirk Weissenborn, Xiaohua Zhai, Thomas Unterthiner, Mostafa Dehghani, Matthias Minderer, Georg Heigold, Sylvain Gelly, et~al.
\newblock An image is worth 16x16 words: Transformers for image recognition at scale.
\newblock In \emph{ICLR}, 2020.

\bibitem[Dubey et~al.(2024)Dubey, Jauhri, Pandey, Kadian, Al-Dahle, Letman, Mathur, Schelten, Yang, Fan, et~al.]{dubey2024llama3}
Abhimanyu Dubey, Abhinav Jauhri, Abhinav Pandey, Abhishek Kadian, Ahmad Al-Dahle, Aiesha Letman, Akhil Mathur, Alan Schelten, Amy Yang, Angela Fan, et~al.
\newblock The llama 3 herd of models.
\newblock \emph{arXiv}, 2407.21783, 2024.

\bibitem[Esser et~al.(2021)Esser, Rombach, and Ommer]{esser2021taming}
Patrick Esser, Robin Rombach, and Bjorn Ommer.
\newblock Taming transformers for high-resolution image synthesis.
\newblock In \emph{CVPR}, pages 12873--12883, 2021.

\bibitem[Huang et~al.(2024)Huang, Sun, Yang, Lyu, Cao, and Qi]{huang2024sc}
Yi-Hua Huang, Yang-Tian Sun, Ziyi Yang, Xiaoyang Lyu, Yan-Pei Cao, and Xiaojuan Qi.
\newblock Sc-gs: Sparse-controlled gaussian splatting for editable dynamic scenes.
\newblock In \emph{Proceedings of the IEEE/CVF Conference on Computer Vision and Pattern Recognition}, pages 4220--4230, 2024.

\bibitem[Jiang et~al.(2024)Jiang, Tu, Liu, Gao, Long, Wang, and Ma]{jiang2024gaussianshader}
Yingwenqi Jiang, Jiadong Tu, Yuan Liu, Xifeng Gao, Xiaoxiao Long, Wenping Wang, and Yuexin Ma.
\newblock Gaussianshader: 3d gaussian splatting with shading functions for reflective surfaces.
\newblock In \emph{CVPR}, pages 5322--5332, 2024.

\bibitem[Kerbl et~al.(2023)Kerbl, Kopanas, Leimk{\"u}hler, and Drettakis]{kerbl20233d}
Bernhard Kerbl, Georgios Kopanas, Thomas Leimk{\"u}hler, and George Drettakis.
\newblock 3d gaussian splatting for real-time radiance field rendering.
\newblock \emph{ACM Trans. Graph.}, 42\penalty0 (4):\penalty0 139--1, 2023.

\bibitem[Kim et~al.(2018)Kim, Goyal, Chawla, Lee, and Kwon]{kim2018attention}
Wonsik Kim, Bhavya Goyal, Kunal Chawla, Jungmin Lee, and Keunjoo Kwon.
\newblock Attention-based ensemble for deep metric learning.
\newblock In \emph{ECCV}, pages 760--777, 2018.

\bibitem[Kingma(2013)]{kingma2013auto}
Diederik~P Kingma.
\newblock Auto-encoding variational bayes.
\newblock \emph{arXiv preprint arXiv:1312.6114}, 2013.

\bibitem[Krizhevsky et~al.(2009)Krizhevsky, Hinton, et~al.]{krizhevsky2009learning}
Alex Krizhevsky, Geoffrey Hinton, et~al.
\newblock Learning multiple layers of features from tiny images.
\newblock 2009.

\bibitem[Lee et~al.(2022)Lee, Kim, Kim, Cho, and Han]{lee2022autoregressive}
Doyup Lee, Chiheon Kim, Saehoon Kim, Minsu Cho, and Wook-Shin Han.
\newblock Autoregressive image generation using residual quantization.
\newblock In \emph{CVPR}, pages 11523--11532, 2022.

\bibitem[Liu et~al.(2024{\natexlab{a}})Liu, Li, Li, and Lee]{liu2024improved}
Haotian Liu, Chunyuan Li, Yuheng Li, and Yong~Jae Lee.
\newblock Improved baselines with visual instruction tuning.
\newblock In \emph{CVPR}, pages 26296--26306, 2024{\natexlab{a}}.

\bibitem[Liu et~al.(2024{\natexlab{b}})Liu, Li, Wu, and Lee]{liu2024visual}
Haotian Liu, Chunyuan Li, Qingyang Wu, and Yong~Jae Lee.
\newblock Visual instruction tuning.
\newblock In \emph{NeurIPS}, 2024{\natexlab{b}}.

\bibitem[Luiten et~al.(2023)Luiten, Kopanas, Leibe, and Ramanan]{luiten2023dynamic}
Jonathon Luiten, Georgios Kopanas, Bastian Leibe, and Deva Ramanan.
\newblock Dynamic 3d gaussians: Tracking by persistent dynamic view synthesis.
\newblock \emph{arXiv preprint arXiv:2308.09713}, 2023.

\bibitem[Luo et~al.(2024)Luo, Shi, Ge, Yang, Wang, and Shan]{luo2024open}
Zhuoyan Luo, Fengyuan Shi, Yixiao Ge, Yujiu Yang, Limin Wang, and Ying Shan.
\newblock Open-magvit2: An open-source project toward democratizing auto-regressive visual generation.
\newblock \emph{arXiv preprint arXiv:2409.04410}, 2024.

\bibitem[Mildenhall et~al.(2021)Mildenhall, Srinivasan, Tancik, Barron, Ramamoorthi, and Ng]{mildenhall2021nerf}
Ben Mildenhall, Pratul~P Srinivasan, Matthew Tancik, Jonathan~T Barron, Ravi Ramamoorthi, and Ren Ng.
\newblock Nerf: Representing scenes as neural radiance fields for view synthesis.
\newblock \emph{Communications of the ACM}, 65\penalty0 (1):\penalty0 99--106, 2021.

\bibitem[Pumarola et~al.(2021)Pumarola, Corona, Pons-Moll, and Moreno-Noguer]{pumarola2021d}
Albert Pumarola, Enric Corona, Gerard Pons-Moll, and Francesc Moreno-Noguer.
\newblock D-nerf: Neural radiance fields for dynamic scenes.
\newblock In \emph{Proceedings of the IEEE/CVF Conference on Computer Vision and Pattern Recognition}, pages 10318--10327, 2021.

\bibitem[Ramesh et~al.(2021)Ramesh, Pavlov, Goh, Gray, Voss, Radford, Chen, and Sutskever]{ramesh2021zero}
Aditya Ramesh, Mikhail Pavlov, Gabriel Goh, Scott Gray, Chelsea Voss, Alec Radford, Mark Chen, and Ilya Sutskever.
\newblock Zero-shot text-to-image generation.
\newblock In \emph{ICML}, pages 8821--8831. Pmlr, 2021.

\bibitem[Razavi et~al.(2019)Razavi, Van~den Oord, and Vinyals]{razavi2019generating}
Ali Razavi, Aaron Van~den Oord, and Oriol Vinyals.
\newblock Generating diverse high-fidelity images with vq-vae-2.
\newblock In \emph{NeurIPS}, 2019.

\bibitem[Rombach et~al.(2022)Rombach, Blattmann, Lorenz, Esser, and Ommer]{rombach2022high}
Robin Rombach, Andreas Blattmann, Dominik Lorenz, Patrick Esser, and Bj{\"o}rn Ommer.
\newblock High-resolution image synthesis with latent diffusion models.
\newblock In \emph{CVPR}, pages 10684--10695, 2022.

\bibitem[Russakovsky et~al.(2015)Russakovsky, Deng, Su, Krause, Satheesh, Ma, Huang, Karpathy, Khosla, Bernstein, et~al.]{russakovsky2015imagenet}
Olga Russakovsky, Jia Deng, Hao Su, Jonathan Krause, Sanjeev Satheesh, Sean Ma, Zhiheng Huang, Andrej Karpathy, Aditya Khosla, Michael Bernstein, et~al.
\newblock Imagenet large scale visual recognition challenge.
\newblock \emph{IJCV}, 115\penalty0 (3):\penalty0 211--252, 2015.

\bibitem[Sun et~al.(2024{\natexlab{a}})Sun, Jiang, Chen, Zhang, Peng, Luo, and Yuan]{sun2024autoregressive}
Peize Sun, Yi Jiang, Shoufa Chen, Shilong Zhang, Bingyue Peng, Ping Luo, and Zehuan Yuan.
\newblock Autoregressive model beats diffusion: Llama for scalable image generation.
\newblock \emph{arXiv preprint arXiv:2406.06525}, 2024{\natexlab{a}}.

\bibitem[Sun et~al.(2024{\natexlab{b}})Sun, Cui, Zhang, Zhang, Yu, Wang, Rao, Liu, Huang, and Wang]{sun2024generative}
Quan Sun, Yufeng Cui, Xiaosong Zhang, Fan Zhang, Qiying Yu, Yueze Wang, Yongming Rao, Jingjing Liu, Tiejun Huang, and Xinlong Wang.
\newblock Generative multimodal models are in-context learners.
\newblock In \emph{CVPR}, pages 14398--14409, 2024{\natexlab{b}}.

\bibitem[Takida et~al.(2022)Takida, Shibuya, Liao, Lai, Ohmura, Uesaka, Murata, Takahashi, Kumakura, and Mitsufuji]{takida2022sq}
Yuhta Takida, Takashi Shibuya, Weihsiang Liao, Chieh-Hsin Lai, Junki Ohmura, Toshimitsu Uesaka, Naoki Murata, Shusuke Takahashi, Toshiyuki Kumakura, and Yuki Mitsufuji.
\newblock Sq-vae: Variational bayes on discrete representation with self-annealed stochastic quantization.
\newblock In \emph{ICML}, pages 20987--21012. PMLR, 2022.

\bibitem[Tian et~al.(2024)Tian, Jiang, Yuan, Peng, and Wang]{tian2024visual}
Keyu Tian, Yi Jiang, Zehuan Yuan, Bingyue Peng, and Liwei Wang.
\newblock Visual autoregressive modeling: Scalable image generation via next-scale prediction.
\newblock \emph{arXiv}, 2404.02905, 2024.

\bibitem[Touvron et~al.(2021)Touvron, Cord, Douze, Massa, Sablayrolles, and J{\'e}gou]{touvron2021training}
Hugo Touvron, Matthieu Cord, Matthijs Douze, Francisco Massa, Alexandre Sablayrolles, and Herv{\'e} J{\'e}gou.
\newblock Training data-efficient image transformers \& distillation through attention.
\newblock In \emph{ICML}, pages 10347--10357, 2021.

\bibitem[Van Den~Oord et~al.(2017)Van Den~Oord, Vinyals, et~al.]{van2017neural}
Aaron Van Den~Oord, Oriol Vinyals, et~al.
\newblock Neural discrete representation learning.
\newblock In \emph{NeurIPS}, 2017.

\bibitem[Villegas et~al.(2022)Villegas, Babaeizadeh, Kindermans, Moraldo, Zhang, Saffar, Castro, Kunze, and Erhan]{villegas2022phenaki}
Ruben Villegas, Mohammad Babaeizadeh, Pieter-Jan Kindermans, Hernan Moraldo, Han Zhang, Mohammad~Taghi Saffar, Santiago Castro, Julius Kunze, and Dumitru Erhan.
\newblock Phenaki: Variable length video generation from open domain textual descriptions.
\newblock In \emph{ICLR}, 2022.

\bibitem[Vinyals et~al.(2016)Vinyals, Blundell, Lillicrap, Wierstra, et~al.]{vinyals2016matching}
Oriol Vinyals, Charles Blundell, Tim Lillicrap, Daan Wierstra, et~al.
\newblock Matching networks for one shot learning.
\newblock In \emph{NeurIPS}, pages 3630--3638, 2016.

\bibitem[Wang et~al.(2024)Wang, Zhang, Luo, Sun, Cui, Wang, Zhang, Wang, Li, Yu, et~al.]{wang2024emu3}
Xinlong Wang, Xiaosong Zhang, Zhengxiong Luo, Quan Sun, Yufeng Cui, Jinsheng Wang, Fan Zhang, Yueze Wang, Zhen Li, Qiying Yu, et~al.
\newblock Emu3: Next-token prediction is all you need.
\newblock \emph{arXiv preprint arXiv:2409.18869}, 2024.

\bibitem[Williams et~al.(2020)Williams, Ringer, Ash, MacLeod, Dougherty, and Hughes]{williams2020hierarchical}
Will Williams, Sam Ringer, Tom Ash, David MacLeod, Jamie Dougherty, and John Hughes.
\newblock Hierarchical quantized autoencoders.
\newblock In \emph{NeurIPS}, pages 4524--4535, 2020.

\bibitem[Xie et~al.(2024)Xie, Zong, Qiu, Li, Feng, Yang, and Jiang]{xie2024physgaussian}
Tianyi Xie, Zeshun Zong, Yuxing Qiu, Xuan Li, Yutao Feng, Yin Yang, and Chenfanfu Jiang.
\newblock Physgaussian: Physics-integrated 3d gaussians for generative dynamics.
\newblock In \emph{CVPR}, pages 4389--4398, 2024.

\bibitem[Yang et~al.(2024{\natexlab{a}})Yang, Yang, Hui, Zheng, Yu, Zhou, Li, Li, Liu, Huang, et~al.]{yang2024qwen2}
An Yang, Baosong Yang, Binyuan Hui, Bo Zheng, Bowen Yu, Chang Zhou, Chengpeng Li, Chengyuan Li, Dayiheng Liu, Fei Huang, et~al.
\newblock Qwen2 technical report.
\newblock \emph{arXiv preprint arXiv:2407.10671}, 2024{\natexlab{a}}.

\bibitem[Yang et~al.(2024{\natexlab{b}})Yang, Gao, Zhou, Jiao, Zhang, and Jin]{yang2024deformable}
Ziyi Yang, Xinyu Gao, Wen Zhou, Shaohui Jiao, Yuqing Zhang, and Xiaogang Jin.
\newblock Deformable 3d gaussians for high-fidelity monocular dynamic scene reconstruction.
\newblock In \emph{CVPR}, pages 20331--20341, 2024{\natexlab{b}}.

\bibitem[Yu et~al.(2021{\natexlab{a}})Yu, Ye, Tancik, and Kanazawa]{yu2021pixelnerf}
Alex Yu, Vickie Ye, Matthew Tancik, and Angjoo Kanazawa.
\newblock pixelnerf: Neural radiance fields from one or few images.
\newblock In \emph{CVPR}, pages 4578--4587, 2021{\natexlab{a}}.

\bibitem[Yu et~al.(2021{\natexlab{b}})Yu, Li, Koh, Zhang, Pang, Qin, Ku, Xu, Baldridge, and Wu]{yu2021vector}
Jiahui Yu, Xin Li, Jing~Yu Koh, Han Zhang, Ruoming Pang, James Qin, Alexander Ku, Yuanzhong Xu, Jason Baldridge, and Yonghui Wu.
\newblock Vector-quantized image modeling with improved vqgan.
\newblock \emph{arXiv preprint arXiv:2110.04627}, 2021{\natexlab{b}}.

\bibitem[Yu et~al.(2023)Yu, Lezama, Gundavarapu, Versari, Sohn, Minnen, Cheng, Birodkar, Gupta, Gu, et~al.]{yu2023language}
Lijun Yu, Jos{\'e} Lezama, Nitesh~B Gundavarapu, Luca Versari, Kihyuk Sohn, David Minnen, Yong Cheng, Vighnesh Birodkar, Agrim Gupta, Xiuye Gu, et~al.
\newblock Language model beats diffusion--tokenizer is key to visual generation.
\newblock \emph{arXiv preprint arXiv:2310.05737}, 2023.

\bibitem[Zhang et~al.(2024{\natexlab{a}})Zhang, Ge, Xu, He, Wang, Qin, Lu, Geng, and Zhang]{zhang2024gaussianimage}
Xinjie Zhang, Xingtong Ge, Tongda Xu, Dailan He, Yan Wang, Hongwei Qin, Guo Lu, Jing Geng, and Jun Zhang.
\newblock Gaussianimage: 1000 fps image representation and compression by 2d gaussian splatting.
\newblock In \emph{ECCV}, pages 327--345. Springer, 2024{\natexlab{a}}.

\bibitem[Zhang et~al.(2024{\natexlab{b}})Zhang, Kuznetsov, Jindal, Chen, Sochenov, Kaplanyan, and Sun]{zhang2024image}
Yunxiang Zhang, Alexandr Kuznetsov, Akshay Jindal, Kenneth Chen, Anton Sochenov, Anton Kaplanyan, and Qi Sun.
\newblock Image-gs: Content-adaptive image representation via 2d gaussians.
\newblock \emph{arXiv preprint arXiv:2407.01866}, 2024{\natexlab{b}}.

\bibitem[Zheng and Vedaldi(2023)]{zheng2023online}
Chuanxia Zheng and Andrea Vedaldi.
\newblock Online clustered codebook.
\newblock In \emph{ICCV}, pages 22798--22807, 2023.

\bibitem[Zhu et~al.()Zhu, Su, Lu, Li, Wang, and Dai]{zhudeformable}
Xizhou Zhu, Weijie Su, Lewei Lu, Bin Li, Xiaogang Wang, and Jifeng Dai.
\newblock Deformable detr: Deformable transformers for end-to-end object detection.
\newblock In \emph{ICLR}.

\end{thebibliography}
}

\end{document}